\documentclass[journal, twoside]{IEEEtran}
\let\labelindent\relax
\usepackage{amsmath}
\usepackage{amssymb}
\usepackage{mathtools}
\usepackage{bm}
\usepackage{tabularx}
\usepackage{ulem}
\usepackage{threeparttable}
\usepackage{booktabs}
\usepackage{diagbox}
\usepackage{array}
\usepackage{multirow}
\usepackage[table]{xcolor}
\usepackage{graphicx}
\usepackage{wrapfig}
\usepackage[lined, ruled, linesnumbered, noend]{algorithm2e}
\usepackage[colorlinks=true, allcolors=blue]{hyperref}

\SetCommentSty{mycommfont}
\usepackage{eqlist}
\usepackage{amsfonts}
\usepackage{enumitem}
\usepackage{lipsum}
\usepackage{arydshln}

\usepackage{comment}

\usepackage{tikz}
\usetikzlibrary{fit, shapes.misc}

\setlist{nosep}
\setlist[itemize]{left=0pt, topsep=0pt, itemsep=0pt}

\newif\ifshownotes
\shownotestrue

\usepackage{CJKutf8}
\definecolor{bleudefrance}{rgb}{0.19, 0.55, 0.91}
\definecolor{awesome}{rgb}{1.0, 0.13, 0.32}
\definecolor{darkgreen}{rgb}{0.0, 0.65, 0.0}
\definecolor{babyblue}{rgb}{0.29, 0.75, 0.93}
\definecolor{black}{rgb}{0,0,0}

\newcommand{\limebox}[1]{\setlength{\fboxsep}{0pt}{\colorbox{lime}{#1}}}

\begin{document}
\title{Differentiable Object Pose Connectivity Metrics\\ for Regrasp Sequence Optimization} 

\author{Liang Qin, Weiwei Wan$^{*}$, and Kensuke Harada%
\thanks{All authors are with the Graduate School of Engineering Science, The University of Osaka, Japan.\\Contact: Weiwei Wan, {\tt\small wan.weiwei.es@osaka-u.ac.jp}.}}
\markboth{Under review by a robotics journal}
{Qin \MakeLowercase{\textit{et al.}}: Analytical Regrasp Planning with Energy-Based Learning: Gradient Optimization over Pre-Sampled Grasp Candidates}
\maketitle


\bstctlcite{IEEEexample:BSTcontrol}

\begin{abstract}
Regrasp planning is often required when one pick-and-place cannot transfer an object from an initial pose to a goal pose while maintaining grasp feasibility. The main challenge is to reason about shared-grasp connectivity across intermediate poses, where discrete search becomes brittle. We propose an implicit multi-step regrasp planning framework based on differentiable pose sequence connectivity metrics. We model grasp feasibility under an object pose using an Energy-Based Model (EBM) and leverage energy additivity to construct a continuous energy landscape that measures pose-pair connectivity, enabling gradient-based optimization of intermediate object poses. An adaptive iterative deepening strategy is introduced to determine the minimum number of intermediate steps automatically. Experiments show that the proposed cost formulation provides smooth and informative gradients, improving planning robustness over other alternatives. They also demonstrate generalization to unseen grasp poses and cross-end-effector transfer, where a model trained with suction constraints can guide parallel gripper grasp manipulation. The multi-step planning results further highlight the effectiveness of adaptive deepening and minimum-step search.
\end{abstract}


\section{Introduction}
\label{sec_introduction}

\begin{figure}[t]
    \centering{\includegraphics[width=1\linewidth]{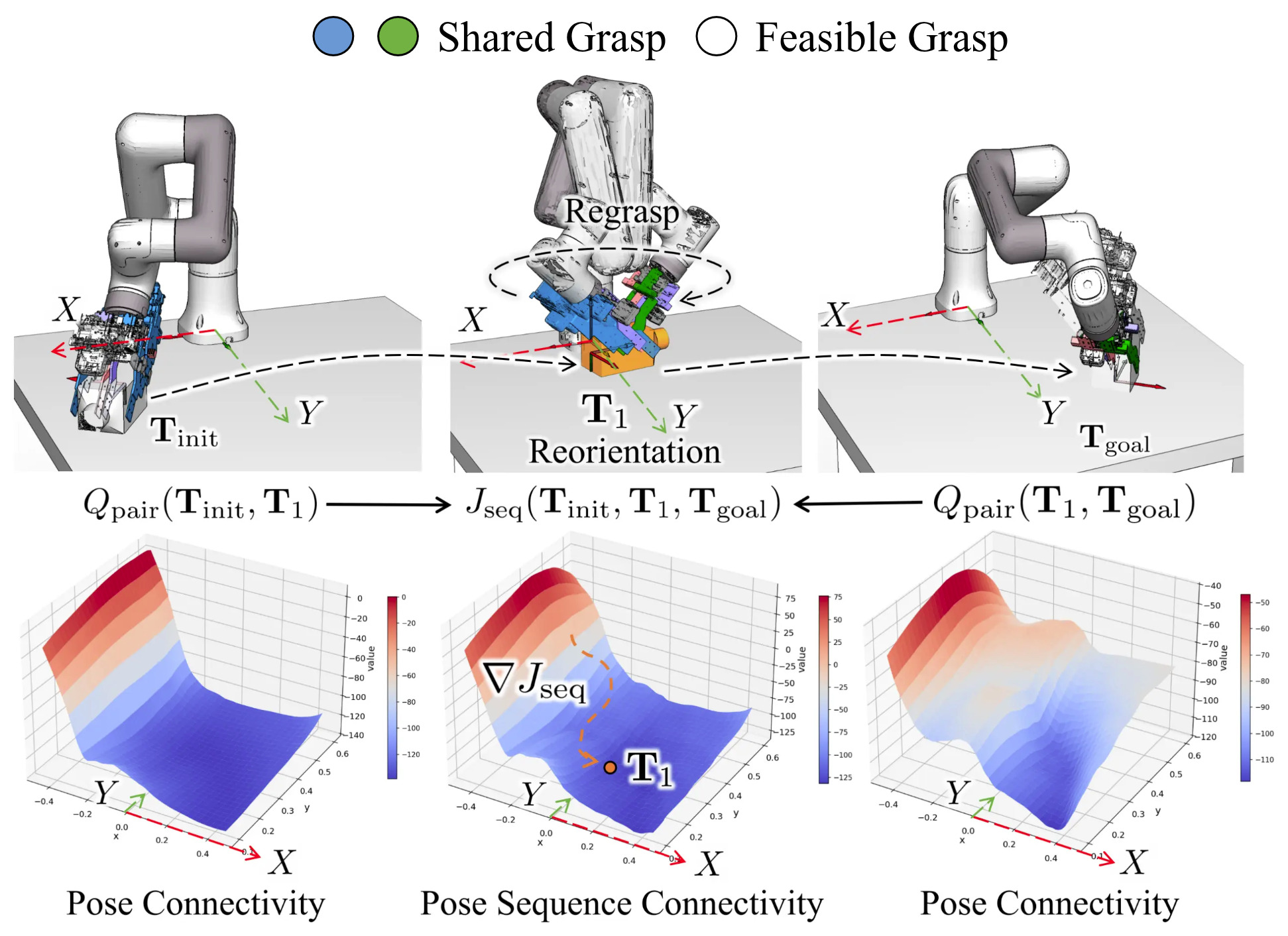}}
    \caption{When one pick-and-place manipulation is infeasible for given object poses $\mathbf{T}_{\mathrm{init}}$ and $\mathbf{T}_{\mathrm{goal}}$, regrasp through one or multiple intermediate poses $\{\mathbf{T}\}_{i=1}^{N}$ (yellow) is required. We propose a \textit{pose connectivity} score $Q_\mathrm{pair}(\mathbf{T}_a, \mathbf{T}_b)$ and a differentiable \textit{pose sequence} cost $J_{\mathrm{seq}}(\mathbf{T}_a,\mathbf{T}_b,\mathbf{T}_c, ...)$ based on EBMs to generate an optimal regrasp sequence by following the gradient $\nabla J_{\mathrm{seq}}$, thus enabling efficient multi-step pick-and-place planning.}
    \label{fig_1}
\end{figure}

\IEEEPARstart{I}{n} robotic pick-and-place, regrasp planning becomes necessary when no grasp can transfer an object from an initial pose to a goal pose under robotic and environmental constraints~\cite{tournassoud1987regrasping}. Classic approaches address this by searching over discretized regrasp graphs constructed from sampled stable placements and grasp candidates~\cite{cho2003complete}\cite{simeon2004manipulation}. However, their performance is fundamentally coupled to sampling density and discretization resolution, and planning can become challenging when the available grasp candidates are highly constrained~\cite{10629240}\cite{raessa2021planning}. In this paper, we propose an implicit planning framework that replaces discrete connectivity search with differentiable connectivity metrics. This framework enables gradient-based intermediate-pose optimization and adaptively determines the minimum number of regrasp steps. (Fig.~\ref{fig_1})

At the core of regrasp planning is \textit{shared grasp}: for each adjacent pose pair in a regrasp sequence, there must exist at least one grasp configuration that is feasible at both poses~\cite{simeon2004manipulation}. When such grasps are missing between the initial and goal poses, intermediate object poses must be introduced to ``bridge'' the sequence, and multiple intermediate steps may be required when feasible grasps are highly constrained.

To address these challenges, we turn shared-grasp constraints into differentiable metrics within an implicit planning framework. Specifically, we model grasp feasibility with an Energy-Based Model (EBM) and derive a \textit{pose connectivity} score by marginalizing composed grasp energies for a pose pair. This yields a smooth scalar field over the object pose space that measures shared-grasp abundance, enabling gradient-based optimization of intermediate poses. We then define a differentiable \textit{pose sequence} cost that aggregates pairwise scores along a chain and includes a regularizer to avoid connectivity bottlenecks. Finally, we combine Langevin dynamics with an adaptive iterative deepening strategy to jointly optimize intermediate poses and search for the minimum required sequence length.

Experiments on a 6-DoF manipulator show that the proposed cost formulation provides smooth and informative gradients, improving planning robustness over alternative metrics. They also demonstrate generalization to unseen grasp poses in one-step intermediate-pose generation and cross-end-effector transferability, where a model trained with suction constraints can guide parallel gripper manipulation. The supplementary video\protect\footnotemark[1]\ published together with this paper shows several real-world execution results with random init and goal poses.

In summary, our key contribution is a differentiable pose sequence connectivity cost derived from compositional EBMs, which transforms discrete shared-grasp constraints into a smooth energy landscape and enables gradient-based optimization of intermediate object poses.

\section{Related Work}
\subsection{Graph-Based Regrasp Planning}

Graph-based approaches construct a connectivity graph from discrete object-grasp pairs, linked by shared poses (transit) or shared grasps (transfer). Planning then becomes a graph search for feasible paths connecting the initial and target poses. Wan et al.~\cite{8611208} applied this approach to single-arm regrasp and dual-arm handover planning in tabletop settings. Addressing environmental complexity, Ma et al.~\cite{8453906} extended this framework to unstructured and cluttered static environments. Beyond standard pick-and-place tasks, graph search methods have been widely applied to dexterous manipulation scenarios. Cruciani et al.~\cite{8594303} introduced graph search to in-hand manipulation, addressing regrasp planning where fingers cannot maintain continuous contact. Building on this, Nagahama et al.~\cite{11037521} utilized graph search to mitigate uncertainty in in-hand manipulation, significantly enhancing grasp precision.
Furthermore, to tackle planning challenges involving complex dynamics or long horizons, researchers have introduced additional modalities and abstraction strategies. Considering grasp stability, Hu et al.~\cite{10341842} incorporated sliding actions into the regrasp framework to expose potentially feasible grasps, establishing a multi-modal transition graph planned via Markov Decision Processes (MDPs). In deformable object assembly tasks, Qin et al.~\cite{10093024} proposed constructing a local Voronoi graph around the object's ideal path, treating regrasping as a passive repair mechanism to resolve geometric blockages. For sequential manipulation puzzles involving rigid bodies, Levit et al.~\cite{10610974} proposed reducing planning dimensionality by searching for sequences of easier sub-problems, using heuristics to guide the operation order. Furthermore, Levit and Toussaint~\cite{11247727} introduced Regrasp Maps, which voxelize space and construct a global state abstraction based on grasp signatures, effectively guiding solvers to discover complex regrasp sequences in highly constrained environments. In this work, we explore an alternative formulation based on a differentiable energy-based cost. By modeling grasp feasibility as an EBM, we compose individual constraints into a combinatorial gradient field, which enables explicit computing derivatives with respect to object poses, thus producing a dense, informative signal derived from the learned grasp manifold to help guide optimization into feasible regions. Moreover, our formulation defines a differentiable objective over an entire intermediate pose sequence, and our algorithm supports adaptively selecting the required number of intermediate steps.

\subsection{Learning-based Regrasp Planning}

Learning methods have made significant progress in regrasp generation. For example, Wada et al.~\cite{wada2022reorientbot} learned to generate feasible reorientation waypoints for visual manipulation. Mishra and Chen~\cite{10610749} utilized diffusion models to sample intermediate poses between start and goal configurations. Xu et al.~\cite{9811547} proposed a hierarchical planning framework that guides multi-step reorientation search by learning a path cost estimator. Recent studies have also focused on predicting diverse stable placements directly from point clouds~\cite{10313307} or constructing incremental regrasp graphs via closed-loop perception~\cite{10955269}. Despite their flexibility and improved performance, these contemporary learning-based approaches primarily target at general regrasp motion generation through learned priors. Different from them, our objective is to combine the controllability of analytical modeling with the efficiency, flexibility, and generalization capability of data-driven methods. In our setting, the object model is available. We analytically precompute table-stable placements from the known object geometry and restrict intermediate poses to planar perturbations around these stable placements. For multi-step regrasp sequence planning, our formulation optimizes the entire sequence of intermediate poses through a differentiable energy-based cost, which enables gradient-based refinement of intermediate states and supports automatic determination of the required number of intermediate steps, rather than relying solely on discrete sampling or fixed-length search.

\section{Energy-Based Regrasp Planning}
\label{sec_regrasp_planning}
This section presents our energy-based formulation for multi-step regrasp planning. We first model grasp feasibility and shared grasps using EBM, then derive differentiable pose and sequence connectivity metrics that enable gradient-based optimization of intermediate object poses. Finally, we describe the adaptive iterative deepening search algorithm used to determine the required number of regrasp steps.

\subsection{Shared Grasp}
\label{subsec_shared_grasp_modeling}
\begin{figure}[!htbp]
    \centering
    \includegraphics[width=1\linewidth]{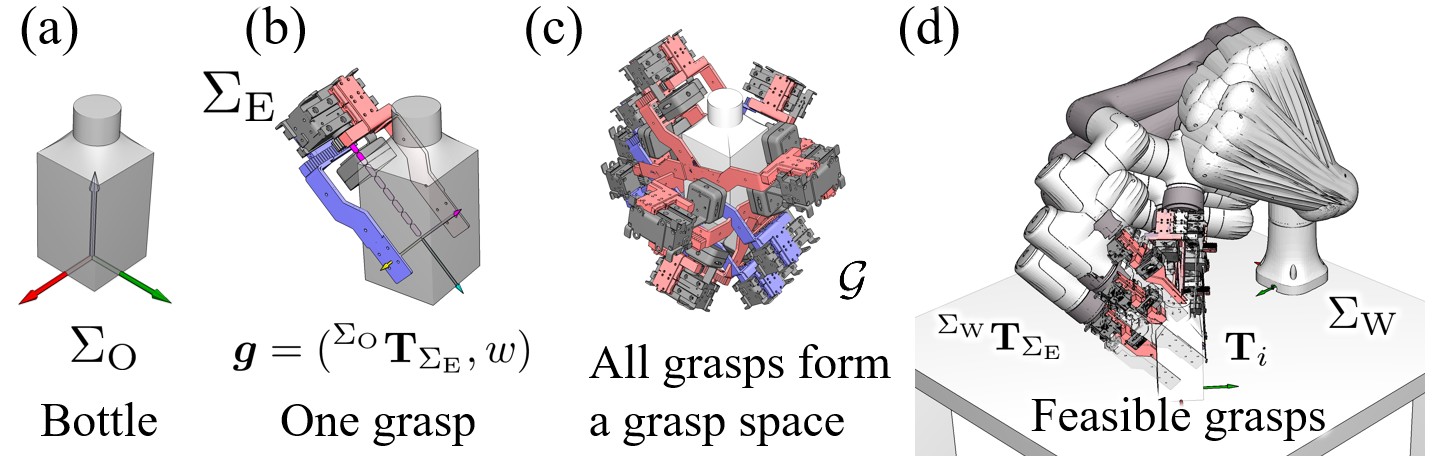}
    \caption{(a) Bottle object. (b) A grasp defined in the object canonical frame. (c) All sampled grasp candidates $\mathcal{G}$. (d) Feasible grasps in the world frame, or namely workspace.}
    \label{fig_feasible_grasp}
\end{figure}

Let $\Sigma_{\mathrm{W}}$, $\Sigma_{\mathrm{O}}$, and $\Sigma_\mathrm{E}$ denote the world frame, the object's canonical frame, and the end-effector frame, respectively. The 6D object pose in the world frame is represented by ${}^{\Sigma_{\mathrm{W}}}\mathbf{T}_{\Sigma_{\mathrm{O}}} \in SE(3) \subset \mathbb{R}^{4 \times 4}$. A grasp configuration is parameterized as a tuple $\boldsymbol{g} = ({}^{\Sigma_{\mathrm{O}}}\mathbf{T}_{\Sigma_\mathrm{E}}, w)$ (Fig.~\ref{fig_feasible_grasp}(a-c)), where ${}^{\Sigma_{\mathrm{O}}}\mathbf{T}_{\Sigma_\mathrm{E}} \in SE(3)$ represents the end-effector pose in the object canonical frame and $w \in [0,1]$ denotes the gripper opening normalized by its maximum stroke. Accordingly, we define the grasp space in the object frame as $\mathcal{G} \triangleq SE(3)\times[0,1]$, with $\boldsymbol{g}\in\mathcal{G}$. Since grasps are represented in $\Sigma_{\mathrm{O}}$, $\mathcal{G}$ is pose-invariant. For any object pose ${}^{\Sigma_{\mathrm{W}}}\mathbf{T}_{\Sigma_{\mathrm{O}}}$ and grasp $\boldsymbol{g} \in \mathcal{G}$, the absolute end-effector pose in the workspace is given by ${}^{\Sigma_{\mathrm{W}}}\mathbf{T}_{\Sigma_\mathrm{E}} = {}^{\Sigma_{\mathrm{W}}}\mathbf{T}_{\Sigma_{\mathrm{O}}} \cdot {}^{\Sigma_{\mathrm{O}}}\mathbf{T}_{\Sigma_\mathrm{E}}$ (Fig.~\ref{fig_feasible_grasp}(d)). For notational simplicity, we omit the frame superscripts/subscripts henceforth, as the reference frames are clear from context.

To evaluate grasp feasibility, we learn an energy function $E_{\phi_f}: SE(3) \times \mathcal{G} \to \mathbb{R}$, where $\phi_f$ denotes the learned parameters of the feasibility energy model. The inputs to the model include the object pose $\mathbf{T}$ and a grasp $\boldsymbol{g}\in\mathcal{G}$, and the output is a scalar energy value. Lower energy values indicate a higher likelihood that $\boldsymbol{g}$ is feasible under $\mathbf{T}$ with respect to IK solvability and collision constraints.

A grasp is defined as a \textit{shared grasp} between an initial object pose $\mathbf{T}_{\mathrm{init}}$ and a target pose $\mathbf{T}_{\mathrm{goal}}$ if the grasp configuration is feasible at both poses~\cite{11247030}\cite{10948354}. We model the prediction of shared grasps as a joint probability estimation problem. Given $\mathbf{T}_{\mathrm{init}}$ and $\mathbf{T}_{\mathrm{goal}}$, we aim to identify grasps $\boldsymbol{g} \in \mathcal{G}$ that lead to a high joint probability $p(\mathbf{T}_{\mathrm{init}}, \mathbf{T}_{\mathrm{goal}}, \boldsymbol{g})$. This joint distribution can be approximated as the product of two independent terms:
\begin{equation}
    p(\mathbf{T}_{\mathrm{init}}, \mathbf{T}_{\mathrm{goal}}, \boldsymbol{g}) \propto p(\mathbf{T}_{\mathrm{init}}, \boldsymbol{g}) \cdot p(\mathbf{T}_{\mathrm{goal}}, \boldsymbol{g}).
    \label{eq_joint_prob}
\end{equation}

Following the Boltzmann distribution $p(\boldsymbol{x}) \propto \exp(-E(\boldsymbol{x}))$, the approximation yields an additive energy form:
\begin{equation}
    p(\mathbf{T}_{\mathrm{init}}, \mathbf{T}_{\mathrm{goal}}, \boldsymbol{g}) \propto \exp\left( - \left[ E_{\phi_f}(\mathbf{T}_{\mathrm{init}}, \boldsymbol{g}) + E_{\phi_f}(\mathbf{T}_{\mathrm{goal}}, \boldsymbol{g}) \right] \right).
    \label{eq_additive_energy}
\end{equation}

This implies that we can efficiently evaluate the joint probability, and thus shared grasps, by summing the energy of a grasp $\boldsymbol{g}$ under $\mathbf{T}_{\mathrm{init}}$ and $\mathbf{T}_{\mathrm{goal}}$~\cite{mitchell2025building,9981264}. 

All shared grasps under given initial and goal poses form a subset $\mathcal{G}_{\mathrm{share}}\subset\mathcal{G}$ where the summed energy falls below a certain threshold $h$~\cite{11316217}:
\begin{equation}
    \mathcal{G}_{\mathrm{share}} = \{ \boldsymbol{g} \in \mathcal{G} \mid E_{\phi_f}(\mathbf{T}_{\mathrm{init}}, \boldsymbol{g}) + E_{\phi_f}(\mathbf{T}_{\mathrm{goal}}, \boldsymbol{g}) \le h \}.
    \label{eq_shared_grasp}
\end{equation}

\begin{figure}[t]
    \centering
    \includegraphics[width=1\linewidth]{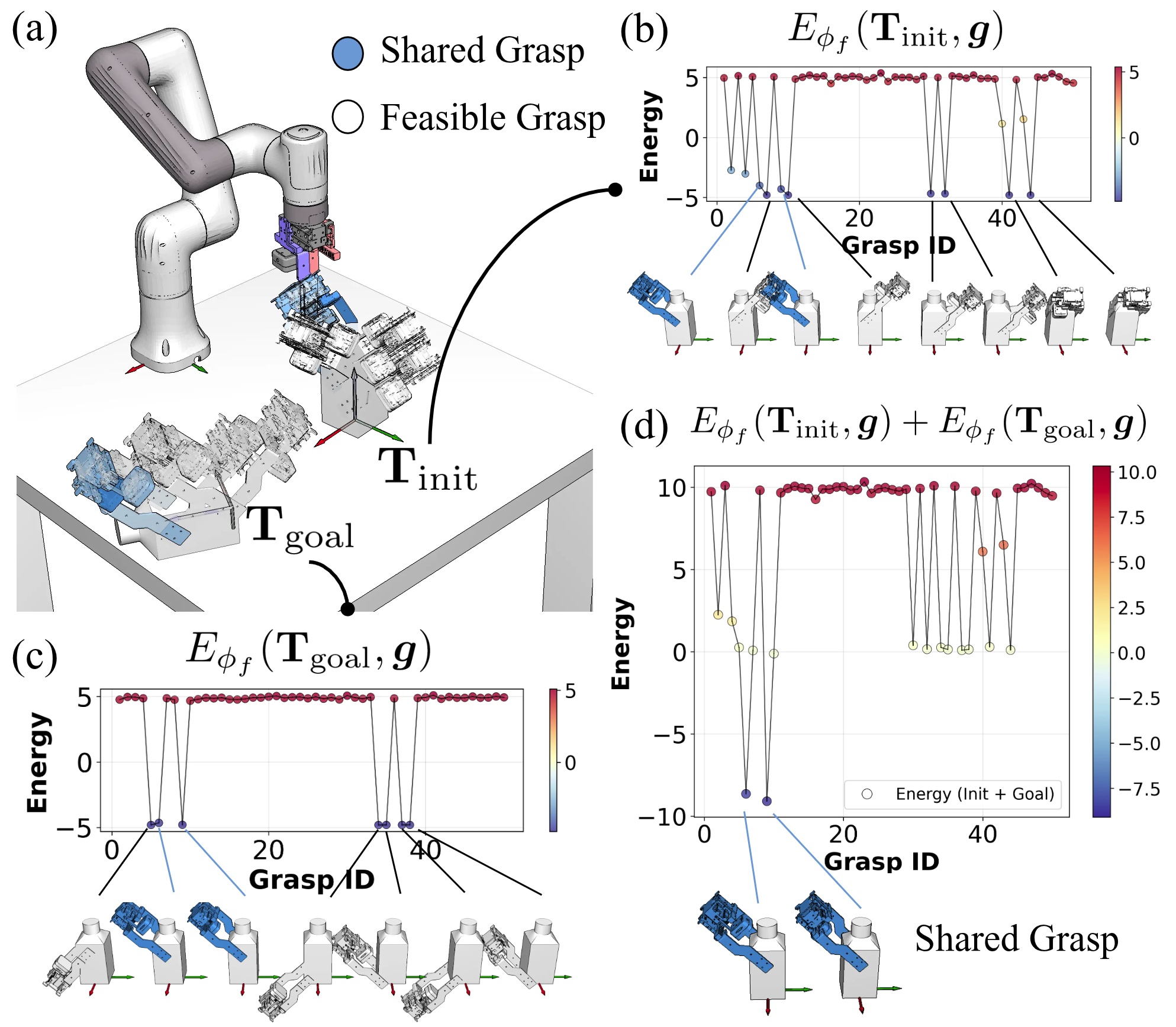}
    \caption{Validation of energy compositionality. (a) Ground truth feasible (white) and shared (blue) grasps. (b)-(c) Individual energy landscapes $E_{\phi_f}$ for initial and goal poses. (d) Joint energy distribution obtained by summation, where low-energy minima correctly align with the shared grasps in (a).}
    \label{fig_energy_composition}
\end{figure}

Fig.~\ref{fig_energy_composition} illustrates this energy additivity: individual energy landscapes assign low energy to feasible grasps under each pose, and summing them yields a joint energy whose minima align with grasps feasible in both configurations.

\subsection{Object Pose Connectivity and Sequence Optimization}
Equation~\eqref{eq_shared_grasp} provides a set-valued characterization of shared grasps. When $\mathcal{G}_{\mathrm{share}}=\emptyset$, a direct transfer from $\mathbf{T}_{\mathrm{init}}$ to $\mathbf{T}_{\mathrm{goal}}$ under one grasp becomes infeasible, and regrasping via intermediate object poses is required. To enable gradient-based search in the object pose space, we define a scalar score to quantify the shared-grasp connectivity between two poses. Specifically, we use a \textit{pose connectivity} score function $Q_\mathrm{pair}(\mathbf{T}_a,\mathbf{T}_b): SE(3)\times SE(3) \rightarrow \mathbb{R}$ to reflect the existence and abundance of shared grasps. Following equation~\eqref{eq_additive_energy}, $Q_\mathrm{pair}(\mathbf{T}_a,\mathbf{T}_b)$ can be approximated by marginalizing the additive grasp energy over the grasp space:
\begin{align}
    &Q_\mathrm{pair}(\mathbf{T}_a, \mathbf{T}_b)\nonumber\\
    &= -\alpha \log \int_{\boldsymbol{g}\in\mathcal{G}} \exp \left( -\frac{E_{\phi_f}(\mathbf{T}_a, \boldsymbol{g}) + E_{\phi_f}(\mathbf{T}_b, \boldsymbol{g})}{\alpha} \right) \, d\boldsymbol{g} \notag \\
    &\approx -\alpha \log \sum_{\boldsymbol{g} \in \mathcal{G}} \exp \left( -\frac{E_{\phi_f}(\mathbf{T}_a, \boldsymbol{g}) + E_{\phi_f}(\mathbf{T}_b, \boldsymbol{g})}{\alpha} \right),
    \label{eq_free_energy_link}
\end{align}
where $\alpha$ is the scaling factor controlling distribution smoothness. In practice, we sample a discrete set of candidate grasps from $\mathcal{G}$ and approximate the integral by finite summation.

Using $Q_\mathrm{pair}$, we can evaluate end-to-end regrasp feasibility along an object pose sequence. We define a differentiable \textit{pose sequence} cost for a sequence $\mathbf{T}_\mathrm{init}, \{\mathbf{T}\}_{i=1}^{N}, \mathbf{T}_\mathrm{goal}$:
\begin{align}
    J_\mathrm{seq}(\{\mathbf{T}&\}_{i=0}^{N+1}) = \underbrace{\sum_{i=0}^{N} Q_\mathrm{pair}(\mathbf{T}_i, \mathbf{T}_{i+1})}_{\text{Connectivity Term}} + \notag \\ 
    \lambda_{\mathrm{reg}} & \underbrace{\sum_{i=0}^{N-1} \left( Q_\mathrm{pair}(\mathbf{T}_i, \mathbf{T}_{i+1}) - Q_\mathrm{pair}(\mathbf{T}_{i+1}, \mathbf{T}_{i+2}) \right)^2}_{\text{Regularization Term}},
    \label{eq_objective_function}
\end{align}
with $\mathbf{T}_0\coloneq\mathbf{T}_{\mathrm{init}}$ and $\mathbf{T}_{N+1}\coloneq \mathbf{T}_{\mathrm{goal}}$ for simplification. Here, $\{\mathbf{T}\}_{i=1}^{N}$ denotes the intermediate pose sequence to be optimized. The cost is differentiable with respect to the intermediate poses, enabling gradient-based optimization. The first term of the definition encourages high shared-grasp connectivity across all adjacent pose pairs in the sequence. The second term is a regularizer that promotes uniform connectivity distribution along the sequence, preventing bottlenecks where one transition is significantly more difficult than others. 

When $N=0$, the chain reduces to $(\mathbf{T}_{\mathrm{init}},\mathbf{T}_{\mathrm{goal}})$ and the objective becomes $J_\mathrm{seq}(\emptyset)=Q_\mathrm{pair}(\mathbf{T}_{\mathrm{init}},\mathbf{T}_{\mathrm{goal}})$, corresponding to a direct pick-and-place without intermediate poses. When $N=1$, the chain reduces to $(\mathbf{T}_{\mathrm{init}},\mathbf{T}_{\mathrm{mid}},\mathbf{T}_{\mathrm{goal}})$ and $J_\mathrm{seq}(\mathbf{T}_{\mathrm{init}},\mathbf{T}_{\mathrm{mid}},\mathbf{T}_{\mathrm{goal}})$ corresponds to pick-and-place with one intermediate regrasp.

\subsection{Minimum-Step Regrasp Sequence Search}
\label{subsec_iterative_planning_method}

Our goal is to find the minimum number of intermediate poses $N$ and the corresponding pose sequence such that every adjacent pose pair along the chain admits at least one shared grasp. We achieve this by converting the set-valued shared-grasp constraint into differentiable metrics: the pairwise pose connectivity score $Q_\mathrm{pair}$ evaluates shared-grasp abundance between two poses, and the pose sequence cost $J_\mathrm{seq}$ aggregates these terms over the full chain (equation~\eqref{eq_objective_function}). Since $N$ is unknown, we adopt an adaptive iterative deepening search method (Algorithm~\ref{alg_regrasp_planning}) that increases $N$ from small to large; for each $N$, it optimizes a batch of candidate sequences by minimizing $J_\mathrm{seq}$, then accepts the first $N$ whose best $K_{\mathrm{top}}$ candidate passes the shared-grasp threshold check on every adjacent pose pair. In practice, Step~3 can be accelerated through a batched threshold check, as shown in Algorithm~\ref{alg_batch_verify}.

\begin{algorithm}[t]
    \small
    \caption{Adaptive Iterative Deepening Search}
    \label{alg_regrasp_planning}
    \SetAlgoLined
    
    \For{$N = 1$ \textbf{to} $N_{\mathrm{max}}$}
    { \label{line_horizon_step}
    \tcp{1. Initialization}
    Initialize batch $\boldsymbol{\Psi}_B = \{{\{\mathbf{T}\}_{i=1}^{N}}^{(b)} \}_{b=1}^B$ via equation~\eqref{eq_ensemble_init} \label{line_init} \;
    
     \tcp{2. Langevin Dynamics}
     \For{$k = 1$ \textbf{to} $K_{\mathrm{opt}}$}{ \label{line_opt_process}
    Compute $\nabla_{\{\boldsymbol{\xi}_i\}_{i=1}^{N}} J_\mathrm{seq}$ for all $\{\mathbf{T}\}_{i=1}^{N} \in \boldsymbol{\Psi}_B$ \;
     Update $\{\boldsymbol{\xi}_i\}_{i=1}^{N}$ using equation~\eqref{eq_langevin_update} \label{line_langevin} \;
     }

     \tcp{3. Sequential Threshold Check}
     ${\{\mathbf{T}\}_{i=1}^{N}}^{(*)} \leftarrow \operatorname*{argmin}_{\{\mathbf{T}\}_{i=1}^{N} \in \boldsymbol{\Psi}_B} \big( J_\mathrm{seq}(\{\mathbf{T}\}_{i=1}^{N}) \big)$ \label{line_min_energy} \;
     $\{\mathbf{T}\}_{i=0}^{N+1} \leftarrow \{\mathbf{T}_{\mathrm{init}}, \{\mathbf{T}\}_{i=1}^{N}, \mathbf{T}_{\mathrm{goal}}\}$ \;
     $is\_valid \leftarrow \mathbf{True}$ \;
      \For{$i = 0$ \textbf{to} $N$}{ \label{line_verify}
          Compute shared set:\allowbreak\,$\mathcal{G}_{\mathrm{share}} \leftarrow \{ \boldsymbol{g}\in\nobreak\mathcal{G} \mid\nobreak E_{\phi_f}(\mathbf{T}_i, \boldsymbol{g}) + E_{\phi_f}(\mathbf{T}_{i+1}, \boldsymbol{g}) \le h \}$ \;
        \If{$\mathcal{G}_{\mathrm{share}} = \emptyset$}{
            $is\_valid \leftarrow \mathbf{False}$ \;
            \textbf{break} \;
        }
     }
    
    \If{$is\_valid$}{
     \Return ${\{\mathbf{T}\}_{i=1}^{N}}^{(*)}$ \label{line_success}
     }
    }
    \Return \textbf{Failure} \;
\end{algorithm}

\begin{algorithm}[t]
    \small
    \caption{Batch Alternative of Step~3}
    \label{alg_batch_verify}
    \SetAlgoLined
    
    $\boldsymbol{\Psi}_{\mathrm{top}} \leftarrow \operatorname*{argmin}^{K_{\mathrm{top}}}_{\{\mathbf{T}\}_{i=1}^{N} \in \boldsymbol{\Psi}_B} \big( J_\mathrm{seq}(\{\mathbf{T}\}_{i=1}^{N}) \big)$ \;
    
    \For{$i = 0$ \textbf{to} $N$}{
        Compute shared set:\allowbreak\,$\mathcal{G}_{\mathrm{share}} \leftarrow \{ \boldsymbol{g}\in\nobreak\mathcal{G} \mid\nobreak E_{\phi_f}(\mathbf{T}_i, \boldsymbol{g}) + E_{\phi_f}(\mathbf{T}_{i+1}, \boldsymbol{g}) \le h \}$ \textbf{for all} $\{\mathbf{T}\}_{i=1}^{N} \in \boldsymbol{\Psi}_{\mathrm{top}}$ \;        
        Update candidates:\allowbreak\,$\boldsymbol{\Psi}_{\mathrm{top}} \leftarrow \{ \{\mathbf{T}\}_{i=1}^{N} \in \boldsymbol{\Psi}_{\mathrm{top}} \mid \mathcal{G}_{\mathrm{share}} \neq \emptyset \}$\;
        
        \If{$\boldsymbol{\Psi}_{\mathrm{top}} = \emptyset$}{
            \Return $\emptyset$ \;
        }
    }
    
    \Return $\operatorname*{argmin}_{\{\mathbf{T}\}_{i=1}^{N} \in \boldsymbol{\Psi}_{\mathrm{top}}} \big( J_\mathrm{seq}(\{\mathbf{T}\}_{i=1}^{N}) \big)$ \;
\end{algorithm}

\begin{figure}[t]
    \centering
    \includegraphics[width=1\linewidth]{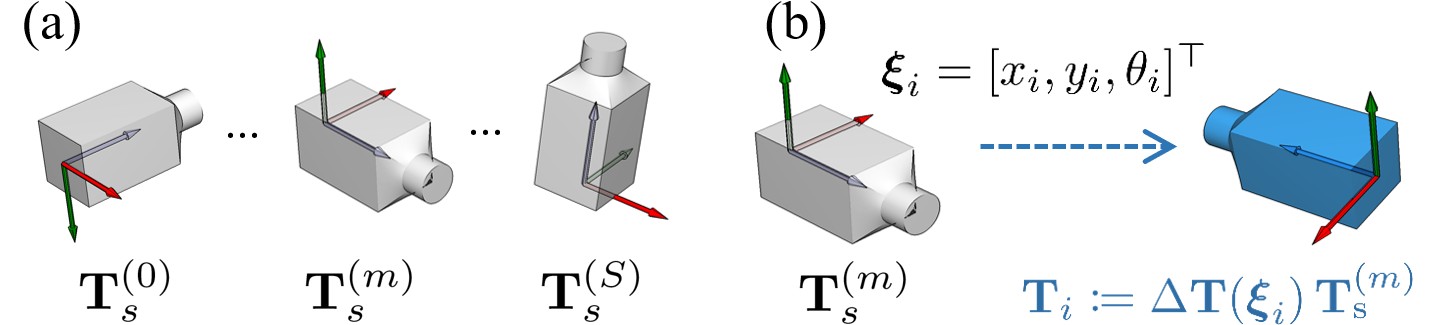}
    \caption{(a) Stable placements on a tabletop environment. (b) Intermediate poses constrained to planar perturbations and rotations.}
    \label{ti}
\end{figure}

In particular, we assume the object has a known shape and is placed in a tabletop environment. Intermediate poses are constrained to planar perturbations around canonical stable placements $\{\mathbf{T}_{\mathrm{s}}^{(m)}\}_{m=1}^{S}$. Each intermediate pose $\mathbf{T}_i$ (with $i=1,\dots,N$) is parameterized by selecting a stable placement index $m \in \{1,\dots,S\}$ and applying planar parameters $\boldsymbol{\xi}_i=[x_i,y_i,\theta_i]^\top$:
\begin{equation}
    \mathbf{T}_{i}
    =
    \Delta \mathbf{T}(\boldsymbol{\xi}_i)\,\mathbf{T}_{\mathrm{s}}^{(m)},
    \quad
    \Delta \mathbf{T}(\boldsymbol{\xi}_i)
    \coloneqq
    \begin{bmatrix}
    \mathbf{R}_z(\theta_i) & [x_i, y_i, 0]^\top \\
    \boldsymbol{0}^\top & 1
    \end{bmatrix},
    \label{eq_ensemble_init}
\end{equation}
where $\mathbf{R}_z(\theta_i)$ denotes a yaw rotation about the world $z$-axis. Fig. \ref{ti} illustrates the constraints. Intermediate sequences are initialized by applying planar perturbations to randomly sampled stable placement sequences.

To perform optimization, we generate a batch $\boldsymbol{\Psi}_B$ containing $B$ sequences of length $N$ using perturbed canonical stable placements. For each particular $N$, we minimize $J_\mathrm{seq}(\{\mathbf{T}\}_{i=0}^{N+1})$ using Langevin dynamics with the following update rule:
\begin{equation}
    \boldsymbol{\xi}_i^{(k+1)} \leftarrow \boldsymbol{\xi}_i^{(k)} - \eta \nabla_{\boldsymbol{\xi}_i} J_\mathrm{seq}({\{\mathbf{T}\}_{i=0}^{N+1}}^{(k)}) + \sqrt{2\eta \tau} \cdot \boldsymbol{z}_k,
    \label{eq_langevin_update}
\end{equation}
where $\boldsymbol{z}_k \sim \mathcal{N}(\boldsymbol{0}, \boldsymbol{I})$ is the random noise term, $\eta$ is the step size, $k$ is the iteration step, and $\tau$ is the temperature parameter. The noise term helps balance local gradient guidance and global exploration for optimized sequence ${\{\mathbf{T}\}_{i=0}^{N+1}}^{(*)}$.

\section{Experiments and Analysis}
We conducted experiments using an Intel Core i9-13900KF (128GB RAM) PC with an RTX 4090 GPU. The data collection, algorithmic implementation, and various studies and verifications are carried out in simulation using the WRS system\footnote{https://github.com/wanweiwei07/wrs}. The real-world deployment are carried out on a 6-DoF Dobot Nova2 manipulator. The workspace was constrained to $x \in [-0.45, 0.45]$ m, $y \in [0.1, 0.6]$ m, and $\theta \in [0, 2\pi]$ rad. The EBM, a three-layer MLP with SELU activations, was trained on 20,000 feasible grasp samples (details in Appendix~\ref{Appendix_EBM_Training}). Optimization used $K_{\mathrm{opt}}=20$ iterations with step size $\eta = 0.3$ and noise $\tau = 0.1$. Connectivity parameters were $\alpha = 1.0$ and $\lambda_{\mathrm{reg}} = 0.5$, with the energy threshold $h$ calibrated individually per object. For fair comparison, we set the batch size to $B=200$ for each stable placement during one-step generation, and scaled it to $B=1000$ random sequences for multi-step planning.

\subsection{Ablation Study of Cost Formulations}

\begin{figure}[!htbp]
    \centering
    \includegraphics[width=1\linewidth]{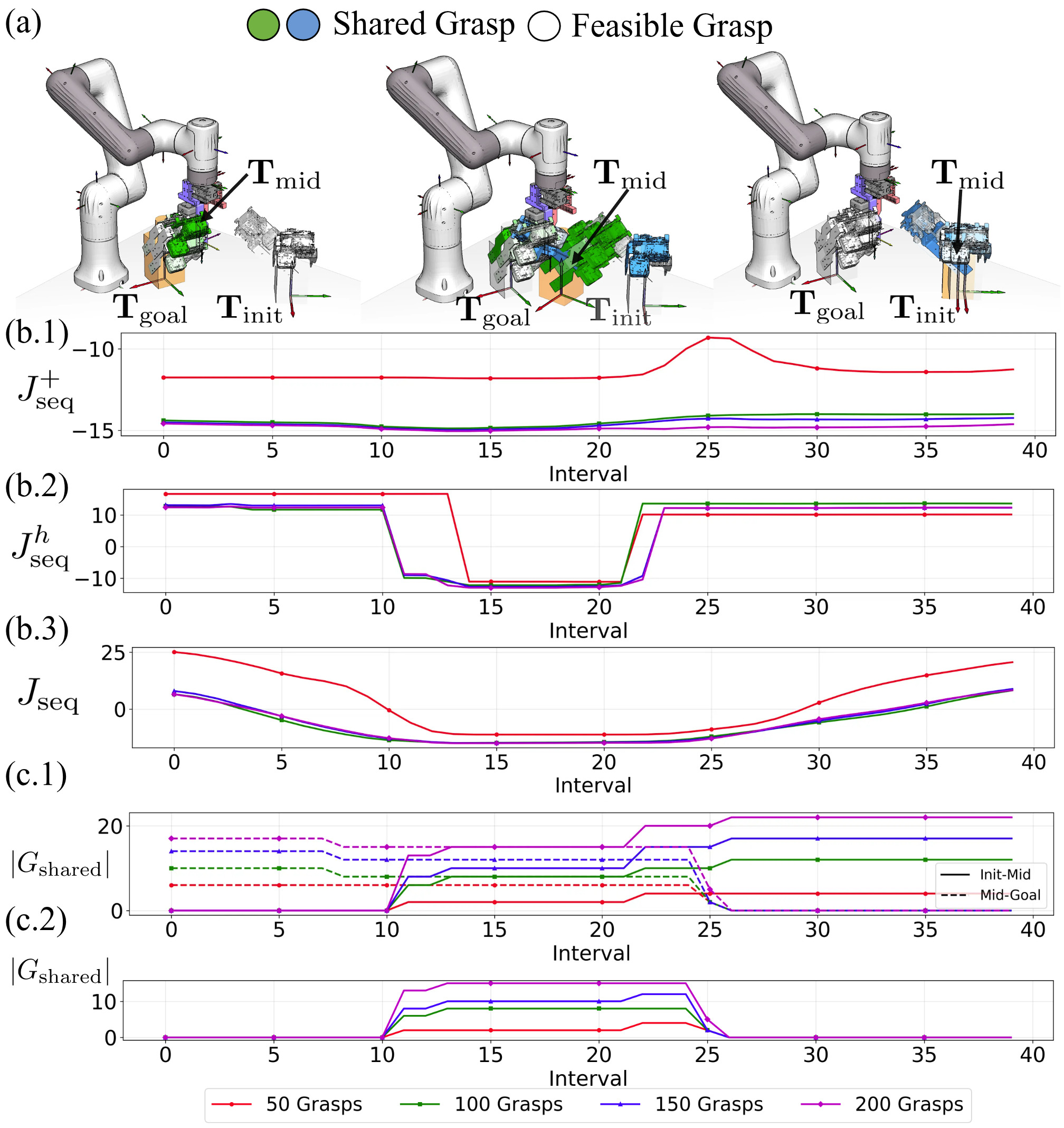}
   \caption{(a) Changes of grasps as $\mathbf{T}_\mathrm{mid}$ shifts from $\mathbf{T}_\mathrm{init}$ to $\mathbf{T}_\mathrm{goal}$. (b) Comparison of different cost formulations. Red, green, blue, and purple curves respectively correspond to results using 50, 100, 150, and 200 sampled candidate grasps from $\mathcal{G}$. (b.1)  $J^{+}_\mathrm{seq}$ leads to spurious minima; (b.2) $J^{h}_\mathrm{seq}$ suffers from vanishing gradients; (b.3) $J_\mathrm{seq}$ ($\lambda_{\mathrm{reg}}=0.5$) has satisfying gradients. (c) Ground truth count of shared grasps (c.1) between adjacent poses, and (c.2) across the whole sequence.}
    \label{fig_shared_grasp_seq_exp}
\end{figure}

We first analyze different pose sequence cost formulations to verify the effectiveness of the proposed $J_\mathrm{seq}$. To isolate the effect of the objective from optimizer stochasticity, we fix a pose pair $(\mathbf{T}_{\mathrm{init}}, \mathbf{T}_{\mathrm{goal}})$ and sweep the intermediate pose $\mathbf{T}_{\mathrm{mid}}$ along a controlled trajectory: a 40-step linear interpolation from $\mathbf{T}_{\mathrm{goal}}$ to $\mathbf{T}_{\mathrm{init}}$ (Fig.~\ref{fig_shared_grasp_seq_exp}(a)). Along this trajectory, we evaluate $J_\mathrm{seq}$ and compare it with other two connectivity formulations: a truncated cost (Appendix~\ref{Appendix_I}):
\begin{equation}
    J^{h}_\mathrm{seq}=Q^{h}_{\mathrm{pair}}(\mathbf{T}_{\mathrm{init}}, \mathbf{T}_{\mathrm{mid}}) + Q^{h}_{\mathrm{pair}}(\mathbf{T}_{\mathrm{mid}},\mathbf{T}_{\mathrm{goal}})+\lambda_{\mathrm{reg}}\mathcal{L}_{\mathrm{reg}},
\end{equation}
and a naive summation cost without the regularizer:
\begin{equation}
    J^{+}_\mathrm{seq}=Q_\mathrm{pair}(\mathbf{T}_{\mathrm{init}}, \mathbf{T}_{\mathrm{mid}}) + Q_\mathrm{pair}(\mathbf{T}_{\mathrm{mid}}, \mathbf{T}_{\mathrm{goal}}).
\end{equation}

Results are shown in Fig.~\ref{fig_shared_grasp_seq_exp}. Fig.~\ref{fig_shared_grasp_seq_exp}(a) illustrates the changes of grasps as $\mathbf{T}_\mathrm{mid}$ shifts from $\mathbf{T}_\mathrm{init}$ to $\mathbf{T}_\mathrm{goal}$. Fig. \ref{fig_shared_grasp_seq_exp}(b.1) reveals the flaw in the naive summation $J^{+}_\mathrm{seq}$: it masks asymmetric connectivity distributions, leading to spurious minima where physical connectivity is broken. Fig. \ref{fig_shared_grasp_seq_exp}(b.2) shows that $J^{h}_\mathrm{seq}$ creates flat potential surfaces in non-connected regions, causing vanishing gradients that hinder optimization. Conversely, Fig. \ref{fig_shared_grasp_seq_exp}(b.iii) demonstrates the necessity of the energy-variance regularizer in $J_\mathrm{seq}$: it eliminates spurious minima and reshapes the energy landscape into a convex structure with significant gradients, ensuring the optimizer effectively converges to a solution where shared grasps exist across the entire $\{{\mathbf{T}_{\mathrm{init}},\mathbf{T}_{\mathrm{mid}},\mathbf{T}_{\mathrm{goal}}\}}$ sequence. Fig.~\ref{fig_shared_grasp_seq_exp}(c) shows the ground truth count of shared grasps for reference. The curves in (c.1) are the shared grasps between two adjacent object poses (\{$\mathbf{T}_{\mathrm{init}},\mathbf{T}_{\mathrm{mid}}$\} or \{$\mathbf{T}_{\mathrm{mid}},\mathbf{T}_{\mathrm{goal}}$\}). The curves in (c.2) are the shared grasps across the whole $\{{\mathbf{T}_{\mathrm{init}},\mathbf{T}_{\mathrm{mid}},\mathbf{T}_{\mathrm{goal}}\}}$ sequence.

\subsection{Comparison with Discrete Search Baseline}
\label{subsub_one_step_generation_exp}

\begin{figure*}[!htbp]
    \centering
    \includegraphics[width=\linewidth]{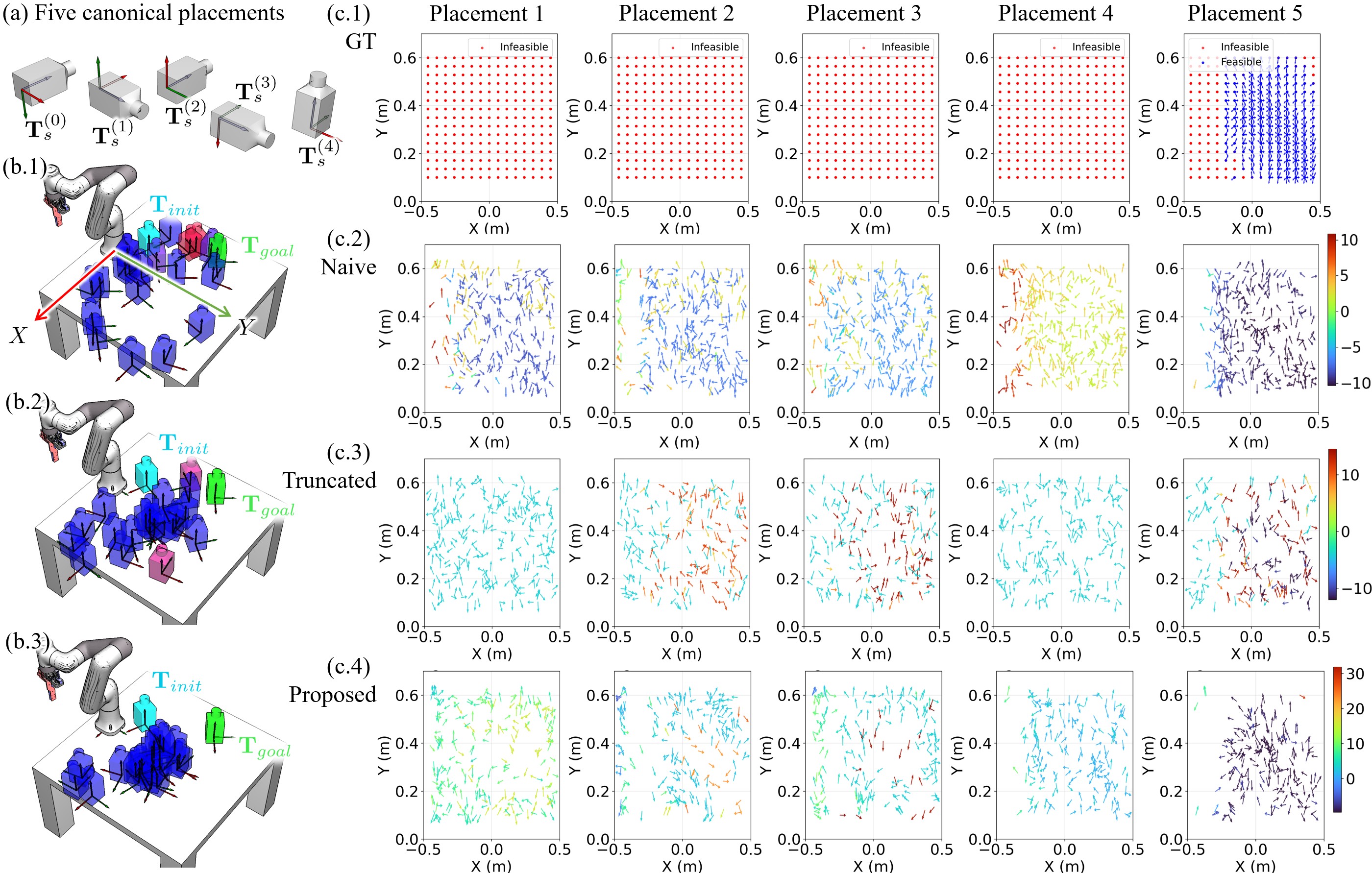}
    \caption{(a) The bottle object has five canonical stable placements. (b.1-3) Evolution of pose distributions and costs during optimization. Since there is only one intermediate pose, we used its color to show the costs. Bluer color denotes lower cost. Redder color denotes a higher cost. (c) Comparison of distributions and costs of the Ground Truth (GT) or generated intermediate poses: (c.1) GT obtained through exhaustive grid sampling. Only perturbations around the canonical stable placement $\mathbf{T}_s^{(4)}$ are valid. Pure red grid points indicate no validity. The points with blue arrows are the valid ones. Each arrow represents a 2D visualization of perturbation $\boldsymbol{\xi}=[x,y,\theta]^\top$. Its starting point is $(x,y)$ and is pointing along the planar yaw $\theta$; (c.2, c.3) Results of the naive cost $J^{+}_{\mathrm{seq}}$ and the truncated cost $J^{h}_{\mathrm{seq}}$. (c.4) Results of the proposed cost $J_\mathrm{seq}$. Similarly, the arrows hold the same representations. The color of the arrows indicates the cost. $J_\mathrm{seq}$ leads to low cost values and similar perturbation distributions for $\mathbf{T}_s^{(4)}$. It also leads to high cost values for $\mathbf{T}_s^{(0)}\sim\mathbf{T}_s^{(3)}$.}
    \label{fig_one_step_exp}
\end{figure*}

\begin{figure}
    \centering
    \includegraphics[width=\linewidth]{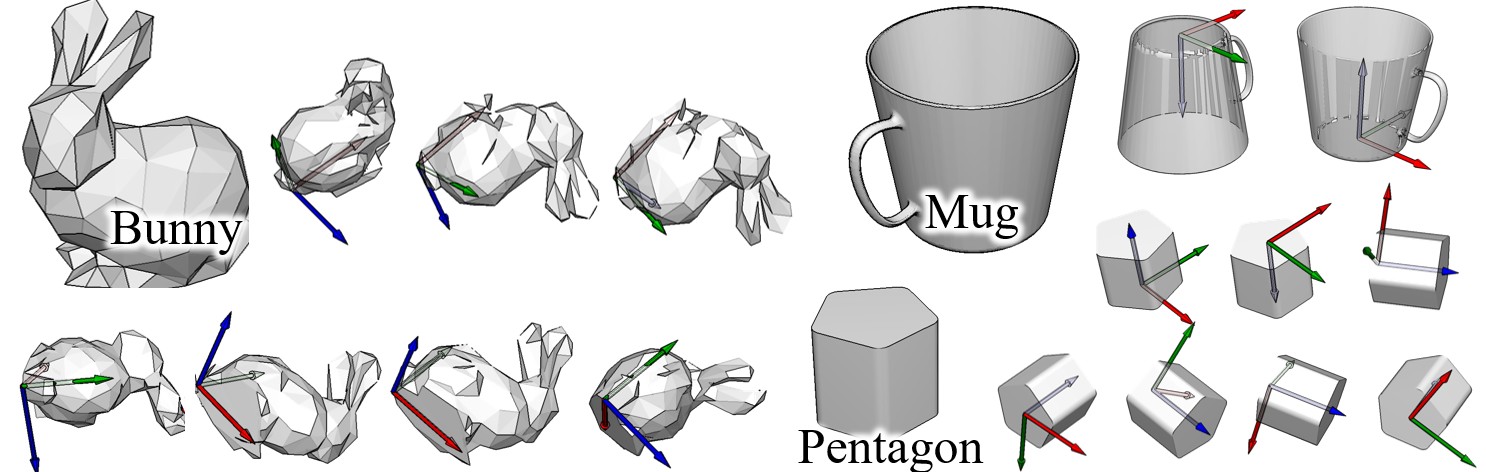}
    \caption{Objects used in the experiments: Bunny, Pentagon, and Mug, in addition to the Bottle shown in Fig.\ref{fig_feasible_grasp}.}
    \label{fig_objs}
\end{figure}

We used a one-intermediate-pose regrasp benchmark to compare $J_\mathrm{seq}$ with a discrete search baseline. The benchmark included four representative objects (Bottle, Bunny, Pentagon, Mug shown in Fig. \ref{fig_feasible_grasp} and Fig. \ref{fig_objs}). We prepared 200 candidate grasps for each object, and tested across 10 sampled $\{\mathbf{T}_{\mathrm{init}}, \mathbf{T}_{\mathrm{goal}}\}$ pairs that lack shared grasps but are guaranteed to be solvable with one intermediate regrasp. The sequence length $N_{\mathrm{max}}$ is thus set to 1 to limit iteration depth.

In the experiments, we used the batched alternative in Algorithm~\ref{alg_batch_verify}. We evaluated $K_{\mathrm{top}}\in\{200,100,50,10\}$ and reported the average verification success rate across all tests. The verification success rate (S.V.) is defined as the fraction of generated intermediate poses that admit IK-feasible and collision-free shared grasps. The discrete search baseline was built based on the canonical stable placements. It established the Ground Truth (GT) intermediate regrasp poses through exhaustive grid sampling on the $SE(2)$ subspace (resolutions: $0.01$ m, $60^\circ$) with IK and collision checking.

Fig. \ref{fig_one_step_exp} visualizes some results of the bottle. The bottle has five canonical stable placements as shown in Fig. \ref{fig_one_step_exp}(a). One optimization process is shown in Fig. \ref{fig_one_step_exp}(b.1-3). Fig. \ref{fig_one_step_exp}(c.4) is the intermediate poses generated by the proposed method. They exhibit a high degree of distribution alignment with the GT in Fig. \ref{fig_one_step_exp}(c.1). Fig. \ref{fig_one_step_exp}(c.2) and (c.3) additionally show the results of the naive and truncated costs for comparison. As expected, they are less effective and deviate from the GT.

Table~\ref{tab_one_step_exp} shows the average S.V. of all tests (the GT verification success rate can be considered as 100\%). The $J_\mathrm{seq}$ cost generally exhibits advantages across all objects. The trend is particularly notable for the Mug, where its complex geometry results in narrow connectivity passages that may pose challenges for $J_\mathrm{seq}^+$ and $J_\mathrm{seq}^h$. Meanwhile, although $J_\mathrm{seq}^+$ and $J_\mathrm{seq}^h$ can be competitive within the narrow $K_\mathrm{top}$=10 range for certain shapes, their performance decreases with larger $K_\mathrm{top}$. This is due to the continuous gradient guidance, which helps reduce the risk of becoming trapped in plateaus or spurious minima.

\begin{table}[t]
\centering
\caption{Average Verification Success Rates (S.V.) of\\ One-Step Intermediate Pose Generation.}
\label{tab_one_step_exp}
\begin{threeparttable}
\setlength\tabcolsep{3.1pt} 
\begin{tabular}{lcccc>{\columncolor{gray!15}}c@{\hspace{12pt}}lcccc>{\columncolor{gray!15}}c}
\toprule
& $K_{\mathrm{top}}$ & $J_\mathrm{seq}$ & $J^{+}_\mathrm{seq}$ & $J^{h}_\mathrm{seq}$ & Usn & & $K_{\mathrm{top}}$ & $J_\mathrm{seq}$ & $J^{+}_\mathrm{seq}$ & $J^{h}_\mathrm{seq}$ & Usn \\
\midrule
\multirow{4}{*}{Bt} & 200 & \limebox{83.1} & 73.0 & 57.1 & 72.9 & \multirow{4}{*}{Bn} & 200 & \limebox{56.0} & 53.0 & 37.2 & 45.8 \\
 & 100 & \limebox{85.9} & 75.3 & 58.0 & 72.8 & & 100 & \limebox{55.3} & 54.6 & 38.3 & 49.8 \\
 & 50 & \limebox{85.4} & 76.6 & 57.6 & 72.2 & & 50 & 53.2 & \limebox{55.6} & 37.2 & 51.6 \\
 & 10 & \limebox{82.0} & 77.0 & 60.0 & 76.0 & & 10 & 48.0 & \limebox{57.0} & 32.0 & 59.0 \\
 \midrule
\multirow{4}{*}{M} & 200 & \limebox{70.8} & 24.9 & 29.9 & 16.1 & \multirow{4}{*}{P} & 200 & \limebox{67.0} & 56.5 & 65.4 & 41.4 \\
 & 100 & \limebox{83.7} & 37.6 & 53.0 & 28.6 & & 100 & \limebox{63.6} & 56.0 & 62.5 & 46.0 \\
 & 50 & \limebox{82.4} & 45.2 & 61.4 & 34.8 & & 50 & \limebox{63.2} & 55.8 & 62.2 & 45.6 \\
 & 10 & \limebox{76.0} & 43.0 & 42.0 & 29.0 & & 10 & 60.0 & 51.0 & \limebox{61.0} & 45.0 \\
\bottomrule
\end{tabular}
\begin{tablenotes}
    \footnotesize
    \item[Note 1] Bt -- Bottle, M -- Mug, Bn -- Bunny, P -- Pentagon.
    \item[Note 2] $J_\mathrm{seq}$ (Proposed) / $J^{+}_\mathrm{seq}$ (Naive) / $J^{h}_\mathrm{seq}$ (Truncated).
    \item[Note 3] Lime cells indicate the highest success rate in each comparison.
    \item[Note 4] The ``Usn'' columns report $J_\mathrm{seq}$ evaluated on unseen grasps.
    \item[Note 5] Values are averaged across 10 $\{{\mathbf{T}_{\mathrm{init}},\mathbf{T}_{\mathrm{mid}},\mathbf{T}_{\mathrm{goal}}\}}$ sequences.
    \end{tablenotes}
\end{threeparttable}
\end{table}

\subsection{Generalization and Transferability}

\subsubsection{Mid-pose generation on unseen grasps}
In this experiment, we evaluated the generalization capability of the $J_\mathrm{seq}$ cost using a separate randomly sampled set of 200 unseen grasps for the same 10 pairs used in experiment~\ref{subsub_one_step_generation_exp}.

As seen in the ``Usn'' columns of Table \ref{tab_one_step_exp}, the method exhibits the capability to generalize to unseen grasp configurations. Although there is an expected performance gap between the ``Usn'' column and the $J_\mathrm{seq}$ column (seen), the gap is acceptable and the S.V. on unseen grasps remains satisfying for less complicated geometries like the Bottle. For the Bunny, the method achieves higher precision in the narrow $K_{\mathrm{top}}$=10 range for unseen grasps. This suggests that the learned energy landscape captures the underlying object geometry effectively, allowing the optimizer to work with the unseen set that were not explicitly covered during training. However, for the Mug and Pentagon, more significant drops are observed. This may stem from: (1) insufficient grasps (200) to capture spatial diversity; (2) unseen grasps inadequately covering successful transitions. Both factors can lead to biased energy evaluations and reduced verification success rate.

\subsubsection{Cross-End-Effector evaluation}
\label{subsec_unseen_end_effectors}
\begin{figure}[!htbp]
    \centering
    \includegraphics[width=1\linewidth]{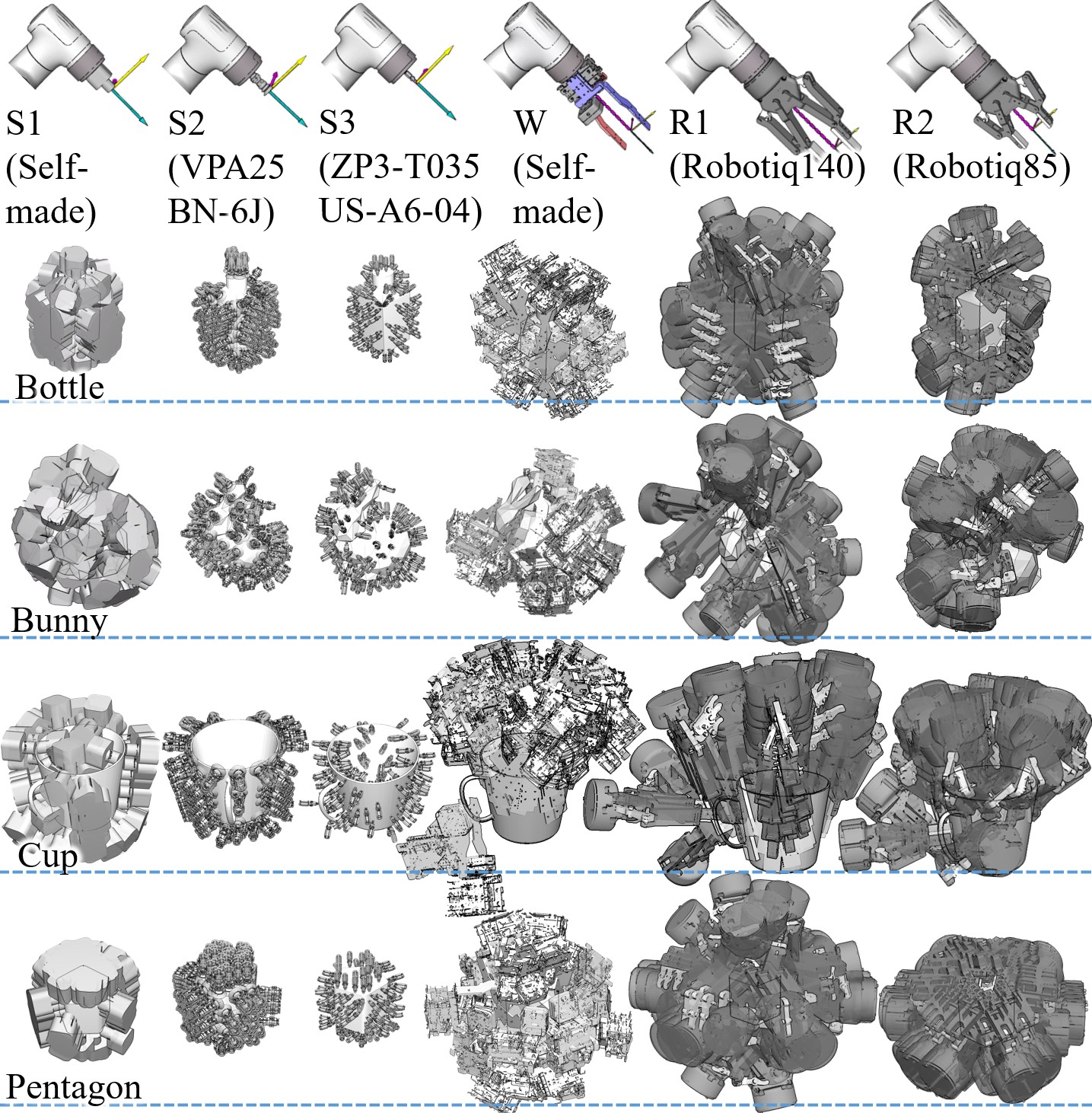}
    \caption{Six end-effectors\protect\footnotemark[2]\ and their candidate grasps used for training.}
    \label{fig_unseen_exp}
\end{figure}

To investigate the versatility of the EBM framework, we evaluated its generalization capability across diverse hardware configurations. Six end-effectors (EEs) were classified based on the strictness of their geometric constraints. We denote the six EEs as W, R1, R2, S1, S2, S3, corresponding to: three parallel grippers (Self-made (W), Robotiq-140 (R1), and Robotiq-85 (R2)), and three suction cups (S1, S2, S3), as shown by Fig. \ref{fig_unseen_exp}. The grippers represent stringent constraints as they require dual-contact alignment and stroke clearance. In contrast, the suction cups represent relaxed constraints and only require single-surface contact. All of them have different Tool Center Points (TCPs). We conducted a $6 \times 6$ cross-evaluation where an EBM trained on one EE was employed to guide one-intermediate pose optimization for another. To ensure a fair comparison, all EBMs were trained on datasets of equal size (20k samples), with the gripper opening width excluded from the input. $K_{\mathrm{top}}$ was set to 100. The reported S.V. was measured using IK and collision checking with the test end-effector. For parallel grippers, collision checking used the actual gripper opening width $w$ associated with each grasp candidate. Evaluation results are summarized in Table~\ref{tab_cross_val_stack}.

\footnotetext[2]{End-effector specifications in meters (TCP offset, maximum opening width): S1 (0.05, $\times$), S2 (0.046, $\times$), S3 (0.03, $\times$), W (0.16, 0.104), R1 (0.20, 0.14), R2 (0.15, 0.085). Suction cups do not have an opening width.}

\begin{table}[!htbp]
    \centering
    \caption{Average Verification Success Rates (S.V.) of\\ Cross-End-Effector Evaluation}
    \label{tab_cross_val_stack}
    \begin{threeparttable}
    \setlength\tabcolsep{1.5pt}
    \begin{tabular}{llcccccc@{\hspace{4pt}}llcccccc}
    \toprule
    & & S1 & S2 & S3 & W & R1 & R2
    & & & S1 & S2 & S3 & W & R1 & R2 \\
    \midrule
    \multirow{6}{*}{Bt}
    & W   & 0.0 & 0.0 & 0.0 & \limebox{86.3} & 67.2 & 63.9
    & \multirow{6}{*}{Bn}
    & W   & 9.2 & 0.0 & 0.0 & \limebox{66.8} & 32.5 & 26.7 \\
    & R1  & 0.0 & 0.0 & 0.0 & 33.0 & \limebox{41.5} & 33.9
    & & R1  & 0.0 & 1.7 & 0.0 & 10.8 & \limebox{35.4} & 0.0 \\
    & R2  & 0.9 & 0.2 & 3.0 & 62.5 & 67.6 & \limebox{64.6}
    & & R2  & 5.5 & 22.9 & 0.0 & 8.1 & 33.7 & \limebox{39.9} \\
    & S1  & \limebox{51.2} & 34.0 & 30.3 & 35.5 & 40.4 & 44.8
    & & S1  & \limebox{39.7} & 29.5 & 1.4 & 25.5 & 4.6 & 0.0 \\
    & S2  & 44.7 & \limebox{40.0} & 36.0 & 42.8 & 33.3 & 49.0
    & & S2  & 4.5 & \limebox{46.1} & 0.6 & 26.2 & 0.1 & 0.0 \\
    & S3  & 34.2 & 20.1 & \limebox{55.5} & 21.3 & 25.6 & 27.2
    & & S3  & 0.0 & 0.0 & \limebox{5.9} & 0.0 & 0.0 & 0.0 \\
    \midrule
    \multirow{6}{*}{M}
    & W   & 0.0 & 11.7 & 11.4 & \limebox{75.7} & 55.7 & 19.7
    & \multirow{6}{*}{P}
    & W   & 17.2 & 9.4 & 9.7 & \limebox{64.2} & 53.9 & 47.7 \\
    & R1  & 0.8 & 7.3 & 5.5 & 51.6 & \limebox{54.3} & 15.9
    & & R1  & 24.6 & 15.6 & 17.0 & 17.2 & \limebox{44.4} & 53.5 \\
    & R2  & 1.5 & 21.0 & 19.5 & 28.9 & 19.6 & \limebox{71.8}
    & & R2  & 18.8 & 8.0 & 16.0 & 16.0 & 32.5 & \limebox{58.1} \\
    & S1  & \limebox{33.6} & 16.6 & 17.9 & 5.7 & 4.9 & 2.6
    & & S1  & \limebox{40.2} & 41.3 & 46.4 & 25.8 & 17.7 & 27.3 \\
    & S2  & 12.3 & \limebox{62.5} & 36.9 & 0.0 & 0.0 & 6.9
    & & S2  & 57.5 & \limebox{57.4} & 54.6 & 41.1 & 27.2 & 41.5 \\
    & S3  & 0.0 & 0.0 & \limebox{20.3} & 0.0 & 0.0 & 0.0
    & & S3  & 62.8 & 69.2 & \limebox{88.1} & 49.4 & 40.8 & 54.5 \\
    \bottomrule
    \end{tabular}
    \begin{tablenotes}
        \item[Note 1] Lime cells indicate the same device for training and testing.
        \item[Note 2] Rows: Training end-effector; Columns: Testing end-effector.
    \end{tablenotes}
    \end{threeparttable}
\end{table}

 The results show: (1) Hardware-specific grasp candidate distributions critically impact performance (e.g., Bottle: W=86.3\%, R1=41.5\%, R2=64.6\%), indicating that end-effector geometry and grasp sampling coverage affect pose connectivity. (2) Cross-modality transfer exhibits asymmetry: suction-trained EBMs can guide gripper-based intermediate-pose optimization with moderate success (Bottle: S1/S2/S3$\rightarrow$W/R1/R2 achieving 21--49\%; Pentagon: 18--55\%), consistent with the intuition that gripper grasp centers form a subset of suction contact distributions for these objects (By 'guide' we mean that the suction-trained EBM is used to optimize intermediate object poses). In contrast, gripper$\rightarrow$suction transfer is much weaker for Bottle (0--3\%) and remains limited for Mug. We attribute this to gripper-specific stroke/clearance constraints and to excluding $w$ in $E_{\phi_f}(\mathbf{T},\boldsymbol{g}')$, while physical verification for parallel grippers still depends on the actual opening width $w$. (3) Optimal end-effector selection is object-specific (e.g., Bottle performs best with W, while Pentagon performs best with S3).

\subsection{Multi-Step Search}
\label{subsec_iterative_planning_exp}

\begin{figure}[!htbp]
    \centering
    \includegraphics[width=1\linewidth]{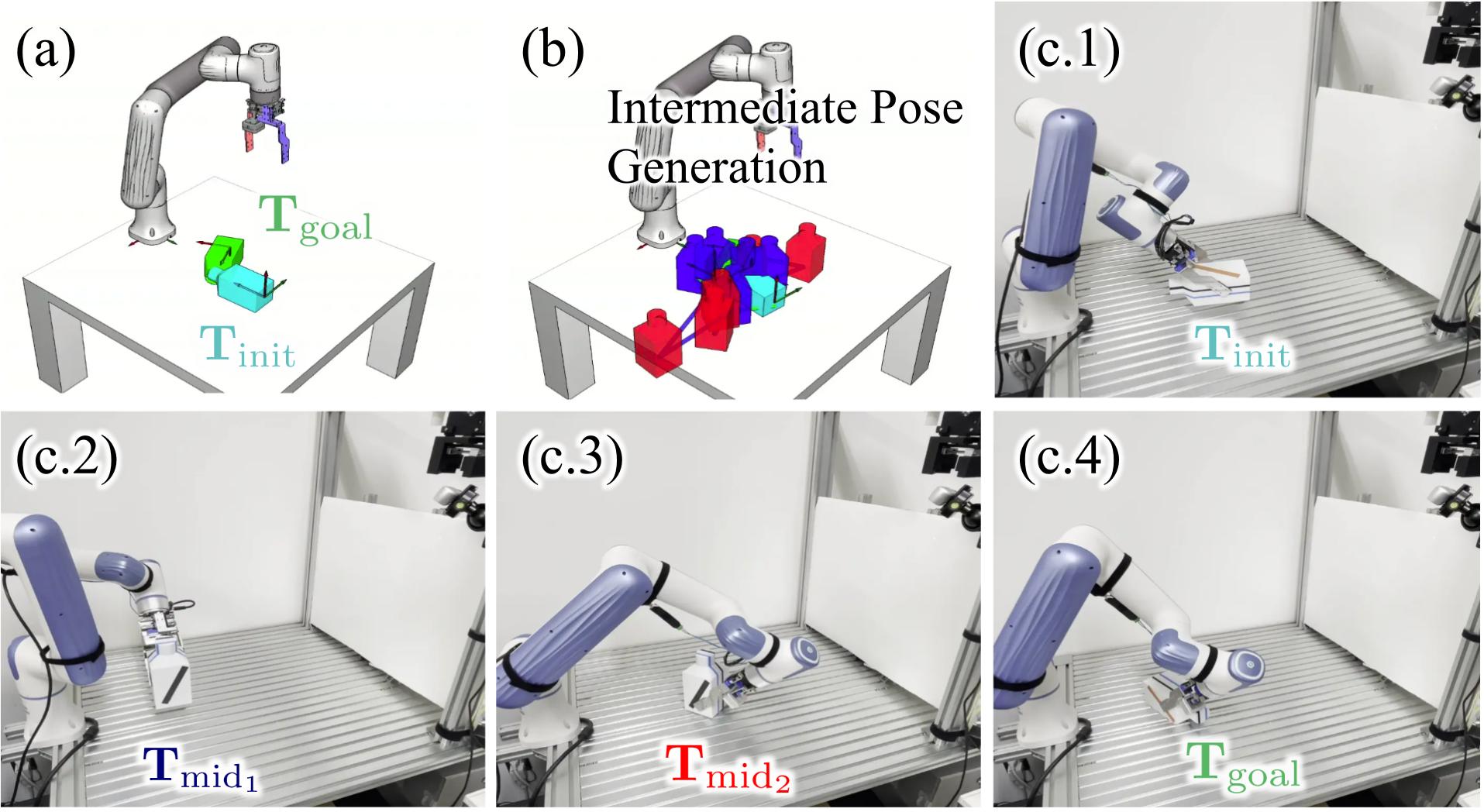}
    \caption{(a) Initial pose $\mathbf{T}_{\mathrm{init}}$ (cyan) and goal pose $\mathbf{T}_{\mathrm{goal}}$ (green). (b) Generation of five candidate intermediate sequences (blue and red). (c.1–4) multi-step pick-and-place motion planning.}
    \label{fig_bottle_multi_step_generation}
\end{figure}

This experiment evaluates the effectiveness of the iterative deepening search. We prepared 50 $(\mathbf{T}_\mathrm{init}, \mathbf{T}_\mathrm{goal})$ pairs that require one or more regrasps. Each episode sampled $B=1000$ intermediate trajectories, with their $K_{\mathrm{top}}$=10 and $K_{\mathrm{top}}$=50 candidates used for batched threshold checking. $N_{\mathrm{max}}$ was set to 5. We report the threshold check success rate (S.T.) and compare it with the hard check baseline (S.H.). The S.T. is defined as the fraction of episodes where the deepening algorithm successfully finds a valid regrasp sequence through threshold checking within $N_{\mathrm{max}}$ steps. The S.H. is defined as the fraction of episodes in which the optimized sequence passes the hard check (i.e., all transitions in the sequence admit at least one shared grasp that is IK-feasible and collision-free). Table \ref{tab_planning_stats_combined} shows the results. It also includes the E.L. (Efficiency Length) and E.T. (Efficiency Time) indices, which report the average sequence length and planning time for successful cases, respectively. Fig.~\ref{fig_bottle_multi_step_generation} exemplifies an optimization process and the deployment of the obtained sequence in the real world.

\begin{table}[!htbp]
    \centering
    \caption{Multi-step search performance.}
    \label{tab_planning_stats_combined}
    \begin{threeparttable}
    \setlength\tabcolsep{3.4pt} 
    \begin{tabular}{lccccc@{\hspace{10pt}}lccccc}
    \toprule
    & $K_\mathrm{top}$ & S.T. & S.H. & E.L. & E.T. & & $K_\mathrm{top}$ & S.T. & S.H. & E.L. & E.T. \\
    \midrule
    \multirow{2}{*}{Bt} & 10 & 42.0 & 48.0 & 1.67 & 1.67 & \multirow{2}{*}{Bn} & 10 & 94.0 & 94.0 & 1.09 & 0.68 \\
    & 50 & 44.0 & 50.0 & 1.82 & 1.97 & & 50 & 94.0 & 96.0 & 1.09 & 0.68 \\
    \midrule
    \multirow{2}{*}{M} & 10 & 10.0 & 10.0 & 1.00 & 0.55 & \multirow{2}{*}{P} & 10 & 90.0 & 84.0 & 1.09 & 0.66 \\
    & 50 & 10.0 & 10.0 & 1.00 & 0.55 & & 50 & 92.0 & 84.0 & 1.17 & 0.83 \\
    \bottomrule
    \end{tabular}
    \begin{tablenotes}
      \footnotesize
      \item[Note 1] S.T. / S.H.: Threshold check success rate. / Hard check success rate.
      \item[Note 2] E.L. / E.T.: Efficiency indices. E.L.: Average sequence length (step count) for successful cases. E.T.: Average planning time (seconds).
      \item[Note 3] Each model was trained on 200 grasp candidates.
    \end{tablenotes}
    \end{threeparttable}
\end{table}


The results demonstrate that iterative deepening effectively restores solvability by adaptively increasing sequence length when needed. The gap between S.T. and S.H. reflects the calibration of the energy threshold $h$: Bottle shows S.H. $>$ S.T., indicating conservative filtering, while Pentagon shows S.T. $>$ S.H., suggesting permissive acceptance. Compared to the Bunny and Pentagon results in Table~\ref{tab_one_step_exp}, multi-step search provides a higher success rate. Bunny and Pentagon also achieve significantly higher success rates (94\% and 90-92\%) compared to Bottle (42-44\%) and Mug (10\%), and less sequence length, which may correlate with their shape-related number of stable placements (7 each vs. 5 and 2) that provide better connectivity options. The minimal performance difference between $K_{\mathrm{top}}$=10 and $K_{\mathrm{top}}$=50 indicates that feasible solutions concentrate in the low-energy region, validating the effectiveness of the energy-based cost formulation. 

\section{Conclusions and Future Work}
In this paper, we introduced differentiable connectivity metrics derived from EBM to enable gradient-based intermediate-pose optimization and developed an iterative deepening search method for regrasp planning. Experiments and analysis validate the effectiveness of the proposed metrics and search framework. Future work will focus on online grasp generation and tighter integration of IK and collision constraints to reduce dependence on pre-sampled grasp sets. 

\normalem
\bibliographystyle{IEEEtran}
\bibliography{citations.bib}

@ARTICLE{11316217,
  author={Anonymous Authors},
  journal={IEEE Robotics and Automation Letters},
  title={Reference Omitted for Double-Anonymous Review},
  year={2025},
  note={Details are withheld to preserve double-anonymous reviewing.}
}

@article{simeon2004manipulation,
  title={Manipulation planning with probabilistic roadmaps},
  author={Sim{\'e}on, Thierry and Laumond, Jean-Paul and Cort{\'e}s, Juan and Sahbani, Anis},
  journal={The International Journal of Robotics Research},
  volume={23},
  number={7-8},
  pages={729--746},
  year={2004}
}

@inproceedings{wada2022reorientbot,
  title={Reorientbot: Learning object reorientation for specific-posed placement},
  author={Wada, Kentaro and James, Stephen and Davison, Andrew J},
  booktitle={International Conference on Robotics and Automation (ICRA)},
  pages={8252--8258},
  year={2022}
}

@INPROCEEDINGS{9981264,
  author={Urain, Julen and Le, An T. and Lambert, Alexander and Chalvatzaki, Georgia and Boots, Byron and Peters, Jan},
  booktitle={IEEE/RSJ International Conference on Intelligent Robots and Systems (IROS)}, 
  title={Learning Implicit Priors for Motion Optimization}, 
  year={2022},
  pages={7672-7679},
  doi={10.1109/IROS47612.2022.9981264}
}

@ARTICLE{8611208,
  author={Wan, Weiwei and Harada, Kensuke and Kanehiro, Fumio},
  journal={IEEE Transactions on Industrial Informatics}, 
  title={Preparatory Manipulation Planning Using Automatically Determined Single and Dual Arm}, 
  year={2020},
  volume={16},
  number={1},
  pages={442-453},
  doi={10.1109/TII.2019.2892772}
}

@ARTICLE{8453906,
  author={Ma, Jiayao and Wan, Weiwei and Harada, Kensuke and Zhu, Qiuguo and Liu, Hong},
  journal={IEEE Transactions on Cognitive and Developmental Systems}, 
  title={Regrasp Planning Using Stable Object Poses Supported by Complex Structures}, 
  year={2019},
  volume={11},
  number={2},
  pages={257-269},
  doi={10.1109/TCDS.2018.2868425}
}

@inproceedings{tournassoud1987regrasping,
  title={Regrasping},
  author={Tournassoud, Pierre and Lozano-P{\'e}rez, Tom{\'a}s and Mazer, Emmanuel},
  booktitle={IEEE International Conference on Robotics and Automation (ICRA)},
  volume={4},
  pages={1924--1928},
  year={1987}
}

@article{cho2003complete,
  title={Complete and rapid regrasp planning with look-up table},
  author={Cho, Kyoungrae and Kim, Munsang and Song, Jae-Bok},
  journal={Journal of Intelligent and Robotic Systems},
  volume={36},
  number={4},
  pages={371--387},
  year={2003}
}

@ARTICLE{10948354,
  author={Liu, Wenhang and Ren, Meng and Song, Kun and Wang, Michael Yu and Xiong, Zhenhua},
  journal={IEEE Robotics and Automation Letters}, 
  title={A Planning Framework for Complex Flipping Manipulation of Multiple Mobile Manipulators}, 
  year={2025},
  volume={10},
  number={5},
  pages={5162-5169},
  doi={10.1109/LRA.2025.3557749}
}

@INPROCEEDINGS{11247030,
  author={Ko, Tianyi and Ikeda, Takuya and Opra, Balázs and Nishiwaki, Koichi},
  booktitle={IEEE/RSJ International Conference on Intelligent Robots and Systems (IROS)}, 
  title={Simultaneous Pick and Place Detection by Combining SE(3) Diffusion Models with Differential Kinematics}, 
  year={2025},
  pages={9425-9432},
  doi={10.1109/IROS60139.2025.11247030}
}

@article{mitchell2025building,
  title={Building Gradient by Gradient: Decentralised Energy Functions for Bimanual Robot Assembly},
  author={Mitchell, Alexander L and Watson, Joe and Posner, Ingmar},
  journal={arXiv preprint arXiv:2510.04696},
  year={2025}
}

@INPROCEEDINGS{10610974,
  author={Levit, Svetlana and Ortiz-Haro, Joaquim and Toussaint, Marc},
  booktitle={IEEE International Conference on Robotics and Automation (ICRA)}, 
  title={Solving Sequential Manipulation Puzzles by Finding Easier Subproblems}, 
  year={2024},
  pages={14924-14930},
  doi={10.1109/ICRA57147.2024.10610974}
}

@INPROCEEDINGS{11247727,
  author={Levit, Svetlana and Toussaint, Marc},
  booktitle={IEEE/RSJ International Conference on Intelligent Robots and Systems (IROS)}, 
  title={Regrasp Maps for Sequential Manipulation Planning}, 
  year={2025},
  pages={17436-17441},
  doi={10.1109/IROS60139.2025.11247727}
}

@INPROCEEDINGS{10341842,
  author={Hu, Jiaming and Tang, Zhao and Christensen, Henrik I.},
  booktitle={IEEE/RSJ International Conference on Intelligent Robots and Systems (IROS)}, 
  title={Multi-Modal Planning on Regrasping for Stable Manipulation}, 
  year={2023},
  pages={10620-10627},
  doi={10.1109/IROS55552.2023.10341842}
}

@INPROCEEDINGS{8594303,
  author={Cruciani, Silvia and Smith, Christian and Kragic, Danica and Hang, Kaiyu},
  booktitle={IEEE/RSJ International Conference on Intelligent Robots and Systems (IROS)}, 
  title={Dexterous Manipulation Graphs}, 
  year={2018},
  pages={2040-2047},
  doi={10.1109/IROS.2018.8594303}
}

@ARTICLE{11037521,
  author={Nagahama, Ryuta and Wan, Weiwei and Hu, Zhengtao and Harada, Kensuke},
  journal={IEEE Robotics and Automation Letters}, 
  title={Bimanual Regrasp Planning and Control for Active Reduction of Object Pose Uncertainty}, 
  year={2025},
  volume={10},
  number={8},
  pages={8027-8034},
  doi={10.1109/LRA.2025.3580327}
}

@ARTICLE{10093024,
  author={Qin, Yili and Escande, Adrien and Kanehiro, Fumio and Yoshida, Eiichi},
  journal={IEEE Robotics and Automation Letters}, 
  title={Dual-Arm Mobile Manipulation Planning of a Long Deformable Object in Industrial Installation}, 
  year={2023},
  volume={8},
  number={5},
  pages={3039-3046},
  doi={10.1109/LRA.2023.3264779}
}

@article{raessa2021planning,
  title={Planning to repose long and heavy objects considering a combination of regrasp and constrained drooping},
  author={Raessa, Mohamed and Wan, Weiwei and Harada, Kensuke},
  journal={Assembly Automation},
  volume={41},
  number={3},
  pages={324--332},
  year={2021},
  publisher={Emerald Publishing Limited}
}

@INPROCEEDINGS{9811547,
  author={Xu, Kechun and Yu, Hongxiang and Huang, Renlang and Guo, Dashun and Wang, Yue and Xiong, Rong},
  booktitle={International Conference on Robotics and Automation (ICRA)}, 
  title={Efficient Object Manipulation to an Arbitrary Goal Pose: Learning-Based Anytime Prioritized Planning}, 
  year={2022},
  pages={7277-7283},
  doi={10.1109/ICRA46639.2022.9811547}
}

@INPROCEEDINGS{10610749,
  author={Mishra, Utkarsh A. and Chen, Yongxin},
  booktitle={IEEE International Conference on Robotics and Automation (ICRA)}, 
  title={ReorientDiff: Diffusion Model based Reorientation for Object Manipulation}, 
  year={2024},
  pages={10867-10873},
  doi={10.1109/ICRA57147.2024.10610749}
}

@ARTICLE{10955269,
  author={Chen, Zhiyuan and Liu, Jiangshan and Chen, Ronghao and Wang, Jiankun},
  journal={IEEE Transactions on Automation Science and Engineering}, 
  title={Closed-Loop Placement Planning for Regrasping and Reconstruction With Single-View RGB-D Images}, 
  year={2025},
  volume={22},
  number={},
  pages={14084-14095},
  doi={10.1109/TASE.2025.3558473}}

@ARTICLE{10313307,
  author={Xu, Peng and Chen, Zhiyuan and Wang, Jiankun and Meng, Max Q.-H.},
  journal={IEEE Transactions on Cognitive and Developmental Systems}, 
  title={Learning to Predict Diverse Stable Placements for Extrinsic Manipulation on a Support Plane}, 
  year={2024},
  volume={16},
  number={3},
  pages={1095-1107},
  doi={10.1109/TCDS.2023.3330989}
}

@ARTICLE{10629240,
  author={Wang, Renpeng and Xie, Yu and Liu, Houde and Zhou, Wei},
  journal={IEEE/ASME Transactions on Mechatronics}, 
  title={Center-of-Mass-Based Object Regrasping: A Reinforcement Learning Approach and the Effects of Perception Modality}, 
  year={2025},
  volume={30},
  number={2},
  pages={1356-1365},
  doi={10.1109/TMECH.2024.3433435}
}

@IEEEtranBSTCTL{IEEEexample:BSTcontrol,
  CTLuse_forced_etal       = "yes",
  CTLmax_names_forced_etal = "3",
  CTLnames_show_etal       = "1",
  CTLdash_repeated_names   = "no",
  CTLname_format_string    = "{f.~}{vv~}{ll}{, jj}",
  CTLname_latex_cmd        = "",
  CTLname_url_prefix       = "[Online]. Available:"
}

\section*{Appendix}
\label{Appendix_I}
\subsection{Truncated Free Energy Score}
We define a truncated pose connectivity score $Q^h_\mathrm{pair}$ by restricting the energy summation to grasps with combined energy below the threshold $h$:
\begin{equation}
\begin{split}
    Q^{h}_{\mathrm{pair}}(\mathbf{T}_a, \mathbf{T}_b) 
    &= -\alpha \log \sum_{\boldsymbol{g} \in \mathcal{G}}
    \mathbb{I}\!\left[
        E_{a}(\boldsymbol{g}) + E_{b}(\boldsymbol{g}) < h
    \right] \\
    &\quad \cdot \exp \left(
        -\frac{E_{a}(\boldsymbol{g}) + E_{b}(\boldsymbol{g})}{\alpha}
    \right).
\end{split}
    \label{eq_truncated_free_energy}
\end{equation}
Here, $E_{a}(\boldsymbol{g}) \!=\! E_{\phi_f}(\mathbf{T}_{a}, \boldsymbol{g})$, $E_{b}(\boldsymbol{g}) \!=\! E_{\phi_f}(\mathbf{T}_{b}, \boldsymbol{g})$, and the indicator function $\mathbb{I}[\cdot]$ ensures that only grasps with combined energy below $h$ contribute to the score. The truncation threshold $h$ is determined statistically. We evaluated the model on a pre-trained shared grasp validation set and selected the energy boundary that maximizes the $F_1$ score for the shared grasp classification task as $h$. When $\mathcal{G}_{\mathrm{share}} = \emptyset$, the absence of feasible samples creates a gradient-free (plateau) state, assigned a constant penalty of 0. This plateau is intrinsic to truncation: excluding sub-optimal grasps removes the directional cues required to guide optimization through infeasible regions, leading to the vanishing gradients shown in Fig. \ref{fig_shared_grasp_seq_exp}(b.2).

\subsection{EBM Training Details}
\label{Appendix_EBM_Training}
We train the energy function $E_{\phi_f}(\mathbf{T}, \boldsymbol{g})$ to assign lower energy to feasible grasps and higher energy to infeasible ones. The training objective combines three loss terms: (1) Negative Log-Likelihood loss ($\mathcal{L}_{\text{nll}}$) that encourages low energy for feasible samples while normalizing via the partition function; (2) Contrastive loss ($\mathcal{L}_{\text{con}}$) that explicitly separates feasible and infeasible grasps; and (3) Regularization loss ($\mathcal{L}_{\text{reg}}$) that prevents energy divergence. The final loss is $\mathcal{L}_{\text{total}} = \mathcal{L}_{\text{nll}} + \mathcal{L}_{\text{con}} + \alpha_{\mathrm{reg}}\mathcal{L}_{\text{reg}}$, where $\alpha_{\mathrm{reg}}$ is a weighting factor~\cite{11316217}.


\end{document}